\documentclass{article}

\usepackage{arxiv}

\usepackage[utf8]{inputenc} 
\usepackage[T1]{fontenc}    
\usepackage{hyperref}       
\usepackage{url}            
\usepackage{booktabs}       
\usepackage{amsfonts}       
\usepackage{nicefrac}       
\usepackage{microtype}      
\usepackage{lipsum}
\usepackage{graphicx}

\usepackage[ruled,vlined]{algorithm2e}
\usepackage{amsmath}
\usepackage{subfigure}
\usepackage{changepage}
\usepackage{comment}

\title{Active Learning of Driving Scenario Trajectories}

\author{
 Sanna Jarl \\
  Department of Computer Science and Engineering \\
   Chalmers University of Technology \\
   Gothenburg, Sweden \\
   \And
 Linus Aronsson \\
  Department of Computer Science and Engineering \\
   Chalmers University of Technology \\
   Gothenburg, Sweden \\
  \And
 Sadegh Rahrovani \\
  Volvo Cars Corporation\\
  Gothenburg, Sweden\\
  \And
   Morteza Haghir Chehreghani \thanks{corresponding author, Email: \texttt{morteza.chehreghani@chalmers.se}}  \\
  Department of Computer Science and Engineering \\
   Chalmers University of Technology \\
   Gothenburg, Sweden \\
}

\begin{document}

\maketitle

\begin{abstract}
Annotated driving scenario trajectories are crucial for verification and validation of autonomous vehicles. However, annotation of such trajectories based only on explicit rules (i.e. knowledge-based methods) may be prone to errors, such as false positive/negative classification of scenarios that lie on the border of two scenario classes, missing unknown scenario classes, or even failing to detect anomalies. On the other hand, verification of labels by annotators is not cost-efficient. For this purpose, active learning (AL) could potentially improve the annotation procedure by including an annotator/expert in an efficient way. In this study, we develop a generic active learning framework to annotate driving trajectory time series data. We first compute an embedding of the trajectories into a latent space in order to extract the temporal nature of the data. Given such an embedding, the framework becomes task agnostic since active learning can be performed using any classification method and any query strategy, regardless of the structure of the original time series data. Furthermore, we utilize our active learning framework to discover unknown driving scenario trajectories. This will ensure that previously unknown trajectory types can be effectively detected and included in the labeled dataset. We evaluate our proposed framework in different settings on novel real-world datasets consisting of driving trajectories collected by Volvo Cars Corporation. We observe that active learning constitutes an effective tool for labelling driving trajectories as well as for detecting unknown classes. Expectedly, the quality of the embedding plays an important role in the success of the proposed framework.
\end{abstract}



\section{Introduction} \label{sec:intro}
New business models and the digitalization of society has revolutionized the automotive sector, like many other industries. Self-driving vehicles are considered to be one of the new trends that is of high importance for the future of the automotive industry with many social and technological impacts. It, alongside the social aspects, could provide promising solutions to reduce accidents in traffic, ease gridlock issues and allow for more comfortable and productive commutes. In order to comfortably integrate Autonomous Drive (AD) in society, it must be safe and tested properly. One important part of ensuring this is collecting and annotating large amounts of driving scenarios, which can then be used for further verification and validation of autonomous drive functionality in virtual and real test environments.

The importance of obtaining a robust and high quality dataset of driving scenarios can be seen in several steps of AD testing and verification \cite{HoseiniRC21}. However, annotation of driving scenario trajectories could become costly and time-consuming. This is due to the annotation task being heavily dependent on human interaction, either to manually label traffic scenario trajectories or verifying the quality of the labels provided by humans. Furthermore, the data concerning autonomous vehicles comes in the form of time series data, containing information about the motions of various road objects, such as cars and other types of vehicles. The complicated nature of this data could increase the risk that an automated annotation algorithm misses rare classes or miss-classifies fringe cases. These are some of the issues that need to be addressed in order to provide scenario catalog/database completeness and also in order to better understand the traffic/driving behavior.


In this paper, we investigate the effectiveness of active learning to annotate AD time series trajectory data. Active learning provides solutions for accurate and robust data labeling at a low cost \cite{CohnGJ96,settles.tr09}. In this paradigm, a small dataset is initially annotated. Then, only the most informative trajectories (data points) are queried to be labeled by an expert/human. One can consider active learning as a sequential decision making procedure wherein at every step, two operations are performed: i) select the next trajectory to be labeled, and ii) update the classification model using the newly labeled trajectory. Its concept is beyond data labeling and annotation; it is studied for example for decision making in troubleshooting as well \cite{ChenRCK17}.

Our contribution is multifold. First, we develop a generic active learning framework for time series data. As will be discussed in the related work section, previous work on active learning for time series tasks is limited because it is specialized to specific types of sequential and temporal data, or assumes a specific classification method \cite{8368343, 9101367, GWEON2021230}. Instead, we separate the modelling of temporal aspects of the data from active learning and classification. This enables us to apply a wide range of different active learning and classification paradigms. To achieve this, we first embed the time series into a latent space in order to extract the temporal and sequential nature of the data. In this work, we study different latent space representations including multivariate Time Series t-Distributed Stochastic Neighbor Embedding (mTSNE) \cite{mtsne2017,van2008visualizing}, Recurrent Auto-Encoder (RAE) \cite{demetriou2020deep} and Variational Recurrent Auto-Encoder (VRAE). To obtain the VRAE embedding, we adapt our developed framework for RAE in \cite{demetriou2020deep} to the variational setting. Given such an embedding, any classification method and active learning query strategy can be used. In this work, we investigate Support Vector Machines (SVM) and Neural Networks (NN) in combination with the query strategies entropy, margin and random sampling.

Second, detecting unknown driving behavior is of high importance for autonomous vehicles. Therefore, we extend the active learning framework to unknown class detection, where the driving trajectories to be queried do not necessarily correspond to an a priori known trajectory type. We achieve this by initially querying known classes, such that the model uncertainty of future unknown classes becomes higher than for the known classes. A reasonable querying strategy should then query the unknown classes at a higher rate. The same embeddings, query strategies and classifiers used for trajectory classification are going to be investigated for this purpose as well.

Third, we investigate novel real-world datasets provided by Volvo Cars Corporation. The datasets consist of information about various driving scenario trajectories which can help advance research on AD.

Finally, we evaluate the performance of the framework (both of active classification and unknown class detection) on the previously mentioned datasets, with promising results.

This paper is an extension of our preliminary work presented in \cite{fastzero2021} wherein we briefly introduced the classification of time series trajectory data using active learning.
In this work, we additionally, i) elaborate further on the proposed framework and discuss its generality and different specifications, ii) perform more investigations on factors that influence the performance of active learning, such as the choice of classifier, class distribution and  the allocated budget, iii) perform additional experimental studies for example on the rate of the improvements by different active learning strategies, and iv) extend the active learning framework to detect unknown classes.

The rest of the paper is organized as follows. In Section 2, we review the related work on active learning. In Section 3, we describe the data and its preparation used in this study. In section 4, we introduce the latent space representations and the embedding methods to be employed within the active learning framework. In Section 5, we describe the framework for active learning, including the classification models and the query strategies. In Section 6, we extend the framework to identify the trajectories with unknown class labels. In Section 7, we perform the experimental studies, and finally, in Section 8, we conclude the paper.

\section{Related Work}
Active learning has been well studied in different contexts and domains, where most of the early work can be found in the survey of \cite{settles.tr09}. It covers, among other methods, uncertainty based query strategies such as entropy and margin sampling, which are utilized in this work.

The study in \cite{Houlsby2011BayesianLearning} uses predictive entropy with the Gaussian Process
Classifier to yield
a Bayesian active learning method called BALD. The method is then investigated for deep learning along with  some approximate methods based on drop-out \cite{GalIG17,BatchBALD19}. Bosser et al. \cite{bosser2020model}  have studied several model-centric and data-centric aspects of active learning with deep neural network models. They show that the margin query strategy yields higher performance compared to alternatives on MNIST and CIFAR-10 datasets, which is consistent with the study in \cite{margin_performs_well}. \cite{AL4Reaction2021} investigates active learning with the margin query strategy for drug discovery in particular for reaction yield prediction in order to identify the successful reactions with a minimal experimental cost. \cite{pimentel2020deep} studies anomaly detection using active learning.

Active learning can be performed in the form of querying class labels or querying pairwise relations. The former is more common \cite{CohnGJ96,GalIG17,Hanneke07}, and  has been widely used in several applications such as robotics, text analysis, image classification, medicine, computer vision, manufacturing and log data analysis \cite{AL4Reaction2021,tong2001active,YanCJ18,Peng_2021_ICCV,8368343}. Querying pairwise relations, on the other hand, has been mainly studied in the context of semi-supervised learning and interactive clustering \cite{NIPS2016_6449,JMLR_Awasthi}, where some works extend the clustering models on positive and negative edge weights such as correlation clustering and shifted min cut \cite{BansalBC04,ChehreghaniICDM17} to (inter)active learning.

Active learning has also been studied for classification of temporal and sequential data. The work in \cite{PengLN17} develops an informativeness measure for  time series classification based on the nearest neighbor classification method, where \cite{8368343} adapts it for battery event detection. Similarly, the works in \cite{9101367,GWEON2021230} study active learning focused on the nearest neighbor paradigm in different settings such as model selection. The Transfer Active Learning (TAL) method \cite{978-3-030-86383-8_48} combines active learning with transfer learning especially for deep learning. The work in \cite{978-3-030-74251-5_33} considers the robustness and quality of assigned labels via a zoom-in step. The safe active learning work in \cite{NEURIPS2018_b197ffde} models time series by a Gaussian process with a nonlinear exogenous input structure while taking some safety constraints into account. The work in \cite{JuniorRM17} studies active learning for animals, vessels, and human movements trajectories, where the extracted features do not represent the full temporal nature of the data.
We note that active learning has been used in other contexts than classification as well, for example sampling the time points over time series \cite{SinghPGBB05}.

However, these methods suffer from several limitations such as tuning critical hyper parameters, being specialized to specific types of sequential and temporal data, or assuming a specific classification method (e.g., the nearest neighbor method). Unlike these task/method-specific methods, we develop a generic active learning framework for temporal data (time series)  that can employ any active learning strategy (informativeness measure) and can be used with any classification method such as Support Vector Machines and Neural Networks. We achieve this by separating the modelling of temporal aspects of the data from active learning and classification.

\section{Data}

The datasets used in this work are provided by Volvo Cars Corporation (VCC). They are in time series format and contain information about objects surrounding the ego car. These objects correspond to other vehicles, which we refer to as target cars. We extract and use the lateral and longitudinal road positions of the surrounding vehicles in order to obtain three different kinds of trajectories, namely \emph{target car left drive by}, \emph{target car right drive by} and \emph{target car cut in}. For simplicity, we drop the word `target' from the scenario names and call them \emph{left drive by}, \emph{right drive by} and \emph{cut in}. Note that the framework we adopt in this study is generic and different types of scenarios could be added for a similar analysis. However, in this work we are particularly interested in AD functionality for congested traffic on the highway. Therefore, the previously mentioned trajectory types are of high importance for the test and release of the AD functionality in our case.

We begin by splitting each of the datasets into three parts: a initially annotated set, a unlabeled set and a test set. These sets can then be used to perform active learning experiments, as explained in Section \ref{sec:active-learning-paradigm}. Then, some of the cut ins are removed to achieve the desired class distribution. The next step is to transform the time series trajectories into a latent space, wherein the temporal aspects are taken into account. After such an embedding, we sometimes call a trajectory a data point, as it is then represented by a vector.  The number of data points (trajectories) in each set for every class distribution can be seen in Table \ref{tab:data_set_size}, where $\alpha$ is the percentage of cut ins. Each set contains an equal number of left and right drive by data points. Finally, it should be mentioned that our datasets were already manually labeled by VCC through expert domain knowledge. These ground truth labels allow us to easily evaluate our active learning framework.


\begin{table}[h]
    \centering
    \begin{tabular}{|c|c|c|c|}
        \hline
        Data set & Annotated set & Unlabelled set & Test set \\\hline
        $\alpha = 33$ & 10 & 2211 & 615 \\
        $\alpha = 10$ & 10 & 1769 & 492 \\
        $\alpha = 5$ & 10 & 1563 & 435 \\\hline
    \end{tabular}
    \caption{The number of data points in the three datasets used in this work, with $\alpha$ being the percentage of cut ins.}
    \label{tab:data_set_size}
\end{table}

\section{Latent Space Representation For Active Learning}
As mentioned, the first step is to model the temporal and sequential order of the trajectories. For this purpose, we embed the trajectories and obtain a data point in a latent space for each trajectory. In this section, we describe the different trajectory embedding methods we investigate in this paper.

\subsection{mTSNE with Dynamic Time Warping}
One way to embed the trajectories into a latent space, is to use mTSNE together with Dynamic Time Warping (DTW) \cite{van2008visualizing,hoseini2020generic}. It first computes the pairwise distances between trajectories using Dynamic Time Warping, and then applies a combination of stochastic neighbor embedding and t-distributed neighbor embedding. To obtain the distance between two trajectories, DTW matches the indices in these time series with some restrictions on the alignment. For example one index in a shorter trajectory might correspond to several indices in a longer trajectory. This is repeated for all pairs of trajectories in order to obtain the matrix of pairwise distances between trajectories (shown by $\mathbf D$).

High dimensional pairwise distances $\mathbf D_{ij}$ are then converted into probabilities of pairwise similarities using an exponential conditional distribution.
The conditional probability

\begin{equation}
    p_{j|i} = \frac{\exp\left( -\mathbf D_{ij}^2/2\sigma_i^2 \right)}{\sum_{k\neq i}\exp\left( -\mathbf D_{ik}^2/2\sigma_i^2 \right)}\,
    \label{eq:SNE}
\end{equation}

is used to imply the probability the $i$-th trajectory (data point) would pick the $j$-th trajectory
as its neighbor if both were drawn in proportion to their probability density under a Gaussian distribution centered around the $i$-th one.
For the low dimensional embedded data points $\mathbf{x}_i$ and $\mathbf{x}_j$
the low dimensional conditional probability $q_{j|i}$ can be computed in a similar way as shown in Eq.  \ref{eq:SNE_embedded}.

\begin{equation}
    q_{j|i} = \frac{\exp\left( -||\mathbf{x}_i - \mathbf{x}_j||^2/2\sigma_i^2 \right)}{\sum_{k\neq i}\exp\left( -||\mathbf{x}_i - \mathbf{x}_k||^2/2\sigma_i^2 \right)}\, ,
    \label{eq:SNE_embedded}
\end{equation}

If the low dimensional data points $\{\mathbf x_i\}$ correctly model the high dimensional distances between trajectories, $p_{j|i}$ and $q_{j|i}$ will be close to each other. Stochastic neighborhood embedding (SNE) aims to find a low dimensional representation (in the form of a vector $\mathbf x_i$ for every $i$-th trajectory) that minimizes the difference between $p_{j|i}$ and $q_{j|i}$.

\begin{figure}[htb]
    \centering
 \subfigure[1024 data points in each class]{\label{subfig:mtsne_1024}\includegraphics[width=0.45\textwidth]{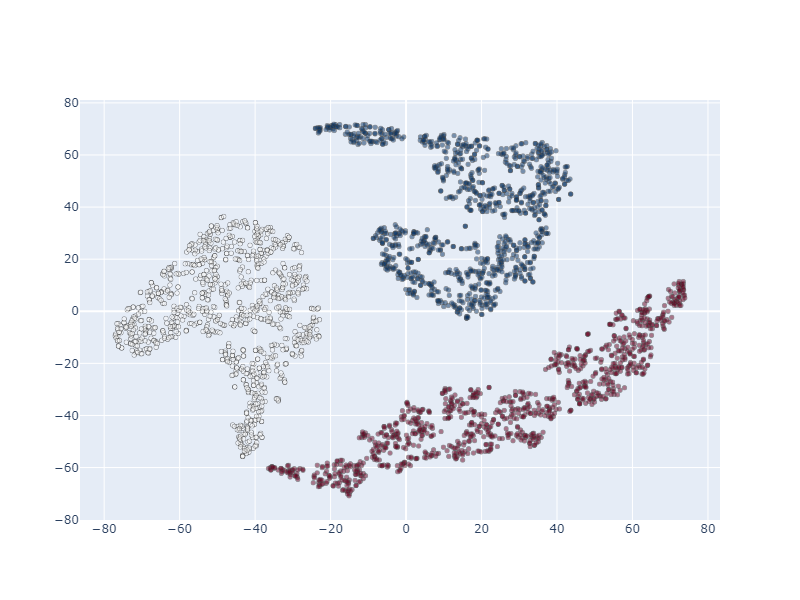}}
    \subfigure[10\% cut ins]{\label{subfig:mtsne_10perc}\includegraphics[width=0.45\textwidth]{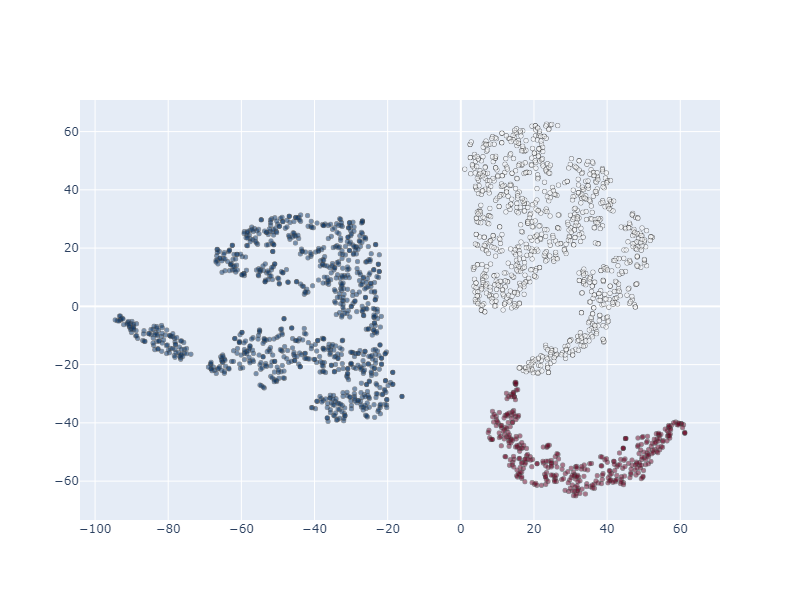}}
    \caption{The latent space representation generated by mTSNE, with (a) 1024 points in each class and (b) 10\% cut ins. The colors red, white and blue correspond to the  cut in, right drive by and left drive by classes, respectively.}
    \label{fig:mtsne_latent}
\end{figure}

To demonstrate how  mTSNE obtains embeddings for our datasets, in Figure \ref{fig:mtsne_latent}, we illustrate  the latent space representation generated by mTSNE, with (a) 1024 data points in each class, and (b) 10\% cut ins. The colors red, white and blue respectively correspond to the cut in, right drive by and left drive by classes. Here we set the perplexity to 37.5, and we see that SNE separates the classes well in two dimensions.

\subsection{RAE and VRAE embeddings}
An alternative approach to produce the latent space representation is using a combination of Recurrent Neural Networks (to extract the temporal aspects of the trajectories) with (Variational) Auto-Encoders (to yield a latent space representation -- an embedding).
We investigate such a representation in two settings: i) Recurrent Auto-Encoder (RAE), and ii) Variational Recurrent Auto-Encoder (VRAE).

\begin{figure}[htb]
    \centering
 \subfigure[RAE]{\label{fig:rae}\includegraphics[width = \textwidth]{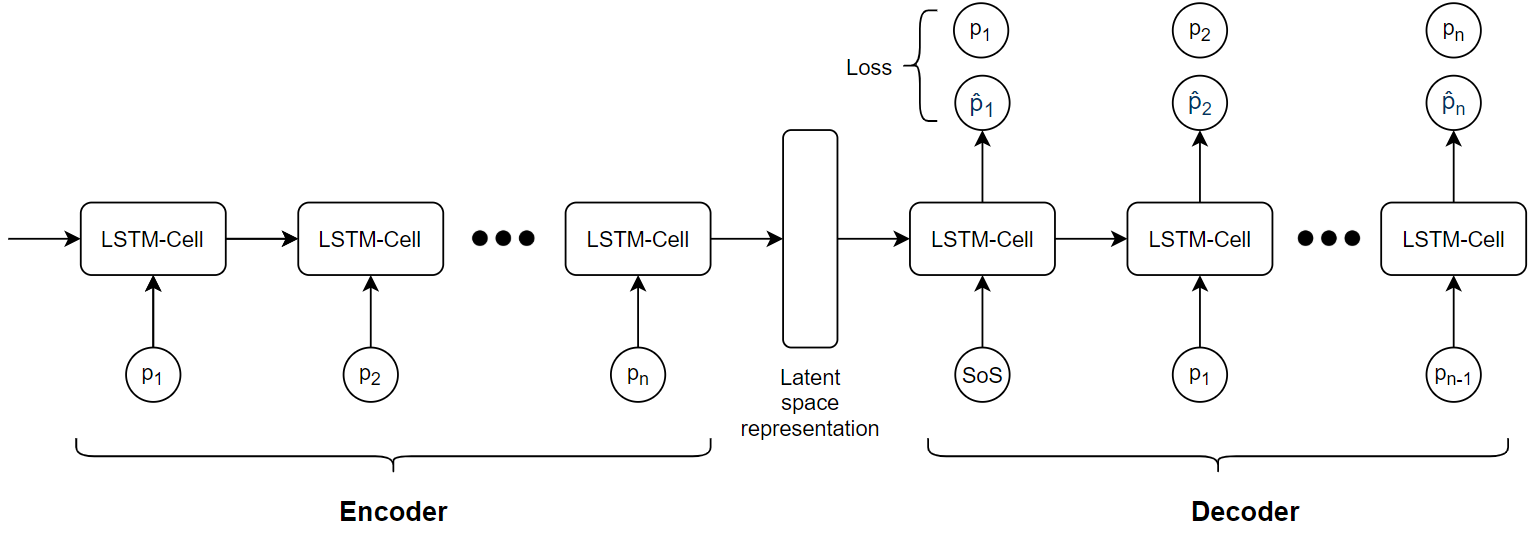}}
     \hfill
    \subfigure[VRAE]{\label{fig:vrae}\includegraphics[width = \textwidth]{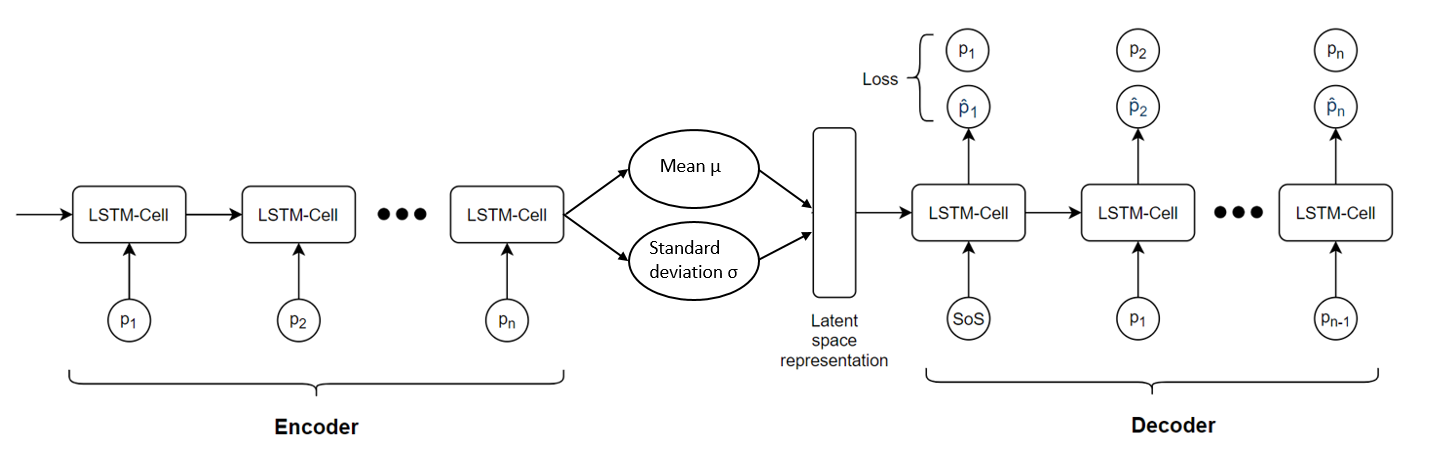}}
   \caption{The general structure of (a) the Recurrent Auto-Encoder (RAE) and (b) the Variational Recurrent Auto-Encoder (VRAE) we use for embeddings.}
    \label{fig:aes}
 \end{figure}

\begin{figure}[htb!]
    \centering

 \subfigure[8 features]{\label{subfig:AE_8}
    \includegraphics[width=0.42\textwidth]{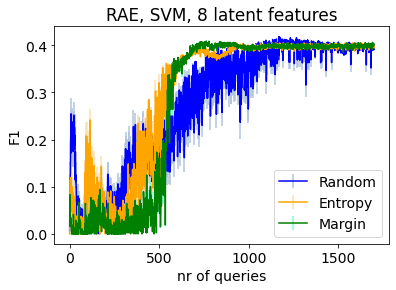}}
    \subfigure[32 features]{\label{subfig:AE_32}
    \includegraphics[width=0.42\textwidth]{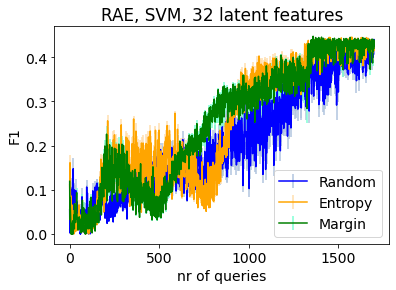}}
    \\
 \subfigure[64 features]{\label{subfig:AE_64}
    \includegraphics[width=0.42\textwidth]{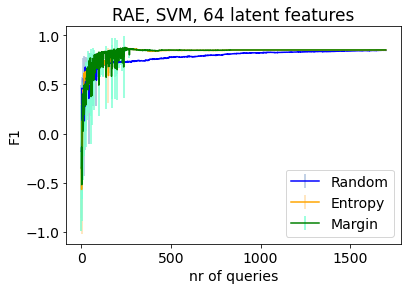}}
    \subfigure[128 features]{\label{subfig:AE_128}
    \includegraphics[width=0.42\textwidth]{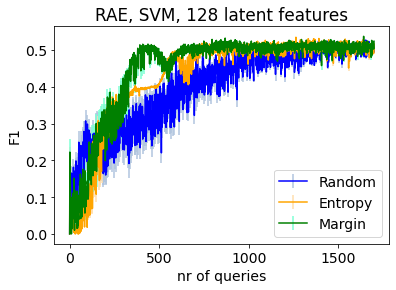}}
  \caption{A comparison of different dimensions of RAE. 64 latent features yields the highest and most stable F1 score.}
    \label{fig:AE_dimensions}
\end{figure}

The schematic structure of RAE is shown in Figure \ref{fig:rae} adapted from the model we have developed in \cite{demetriou2020deep} for trajectory generation. Figure \ref{fig:vrae} shows the general structure of VRAE, that includes an additional layer mapping each trajectory (data point) to a distribution in the latent space. The Auto-Encoder has two stacked LSTM cells and 64 features in the hidden state as well as in the latent space. To verify the performance of the autoencoder, we have performed a small study with 8, 32, 64 and 128 features in latent space. As seen in Figure \ref{fig:AE_dimensions}, 64 features yield the most stable and accurate results. In Figure \ref{fig:aes}, the variables $p_1,\hdots,p_n$ represent a sequence of feature vectors that correspond to a driving trajectory, which is input to the autoencoder. In order to address the variable-length problem with the trajectories in our datasets, we group together the trajectories of a certain length to form a batch, which is then fed as input to the network. In this fashion, all trajectories within a batch have the same length.

VRAE ensures a coherent latent space, which implies the data points in close proximity will be located close to each other even in the latent space. VRAE encodes the data points as distributions rather than explicit points, then a new sample is drawn from these distributions as input to the decoder.

\section{The Active Learning Paradigm} \label{sec:active-learning-paradigm}
The goal of active learning is to label the most informative data points with a minimal number of human interactions. In our framework, the active learning procedure uses the latent representation of the time series data as input. To perform active learning, we follow Algorithm \ref{alg:activelearning}, which begins by training the classification model on a small annotated dataset. Next, we classify the unlabeled data and calculate the informativeness of each data point according to the chosen query strategy. The most informative data point is then queried to be annotated by an expert and is subsequently moved to the annotated set. After every query, the classifier is retrained. This iterative procedure continues until the allocated budget is spent.
\begin{algorithm}
    \SetAlgoLined
    \KwResult{Labels a number of data points and classifies the rest. }
     Train the classifier on a small amount of initially labeled data\;
     \While{budget $>$ 0}{
        Classify the unlabeled data points\;
        Calculate informativeness for each unlabeled data point\;
        Query the most informative data point(s) to an expert\;
        Add queried data point(s) to the annotated data set and remove them from the unlabeled set\;
        Retrain the classifier using the (new) annotated set\;
    }
    \caption{Active Learning}
    \label{alg:activelearning}
\end{algorithm}

\subsection{Classification models}
In this study, we examine two classification models,  SVM and fully connected neural network (NN). For SVM, the radial basis function is used as the kernel. For the parameters $c$ and $\gamma$ we use the values suggested by Scikit-Learn. Performing cross-validation by inspection, these parameter values provided good enough results. The NN consists of two hidden layers with respectively 128 and 256 neurons. VRAE seems to require a larger capacity. Thus, we use a larger NN with that, consisting of 5 hidden layers with 64, 128, 256, 128 and 64 neurons. Each layer is batch normed and ReLU is used as the non-linear activation function. To optimize, we use the Adam optimizer applied to the respective cross entropy loss.

\subsection{Query strategies}
We investigate the three commonly-used query strategies \emph{random}, \emph{margin} and \emph{entropy}. As discussed in \cite{bosser2020model}, they represent different aspects of active learning for neural networks. Let U denote the unlabeled set. Random assigns a uniformly distributed informativeness to each data point in U,

\begin{equation}
    I^R_i \sim \text{unif}(0,1).
    \label{eq:uniform}
\end{equation}

The second strategy, called margin, computes the informativeness for every unlabeled data point $\mathbf{x}_i \in U$ as

\begin{equation}
    I^M_i = -[P_C(\hat{y} = \tilde{c}_1|\mathbf{x}_i) - P_C(\hat{y} = \tilde{c}_2|\mathbf{x}_i)],
    \label{eq:margin}
\end{equation}

where $\mathbf{x}_i$ is the latent space representation of the data point, $C$ represents a classifier, $\hat{y}$ is a random variable that corresponds to the predicted label by the classifier, $P_C$ is the probability of $\hat{y} = \tilde{c}_j$ given $\mathbf{x}_i$ (here $j=1,2$), $\tilde{c}_1$ and $\tilde{c}_2$ are respectively the most probable and the second most probable classes for $\mathbf{x}_i$ to belong to predicted by classifier C. The data point with the smallest margin is then queried.

The third query strategy is entropy, which assigns the informativeness based on the entropy of the predictive distribution:

\begin{equation}
    I^E_i = -\sum_{\tilde c_j} P_C(\hat{y} = \tilde{c}_j|\mathbf{x}_i))\log (P_C(\hat{y} = \tilde{c}_j|\mathbf{x}_i))).
    \label{eq:entropy}
\end{equation}
We note that here the summation is w.r.t. all class labels $\{\tilde c_j\}$.
The entropy can be viewed as the total amount of information in the entire distribution. The data point with the highest $I^E_i$ is queried.

\section{Discovering Unknown Classes}
In this section, we extend the proposed framework to employ active learning for finding unknown classes. To do so, we assume that the classification model (SVM or NN) is only trained based on two classes, whereas there might exist more classes in the set of unlabeled data which belong to an unknown class. Then, the goal is to identify such data points with unknown class labels. This can potentially be used for anomaly detection as well.

For this purpose, we hypothesize that when performing  active learning, after a sufficient number of data points is queried from the existing (known) classes, then a reasonable querying strategy might query mainly from the unknown classes. The reason is that the classification model then becomes confident about the existing classes and yields the highest uncertainty for the data points with unknown class labels.

Here, we treat the cut in class as the unknown class, where the classification models use only the data points labeled as left and right drive by. Thus, the entire model including the embeddings (e.g., the Auto-Encoders) are trained on only left and right drive by data points. We use the number of queried cut ins w.r.t. the total number of queried data points a measure of performance.

\begin{figure}[htb]
    \centering
    \subfigure[Margin]{\label{subfig:queried_trajs_margin}\includegraphics[width=0.45\textwidth]{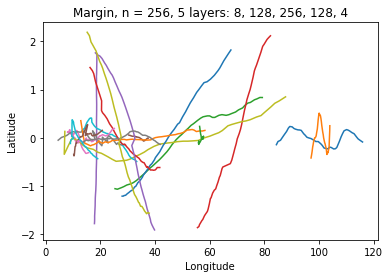}}
    \subfigure[Entropy]{\label{subfig:queried_trajs_entropy}\includegraphics[width=0.45\textwidth]{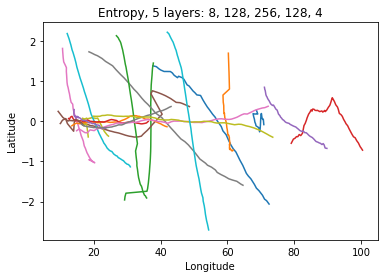}}
    \caption{A subset of queried trajectories using (a) margin and (b) entropy query strategies.}
    \label{fig:queried_trajs}
\end{figure}

To justify our idea of using active learning for unknown class detection, here we carry out a preliminary study to investigate what type of trajectories the active learning method tends to query. Figure \ref{fig:queried_trajs} shows a subset of queried trajectories using the margin (Figure \ref{subfig:queried_trajs_margin}) and entropy (Figure \ref{subfig:queried_trajs_entropy}) strategies (note that in our datasets, the velocity is relative to the ego vehicle). With both query strategies we observe that several double cut ins and decelerative cut ins are queried, which are rare forms of cut in trajectories. This discovery indicates the potential for active learning to be used for the purpose of finding unknown classes or even anomalies of a certain class.

\section{Experimental Studies}
In this section, we investigate the different components of the proposed framework for active learning of the AD trajectories: embedding, query strategy, classifier and class distribution. The embeddings used are mTSNE, RAE and VRAE combined with the SVM or NN classification models. We consider three query strategies, random, margin and entropy. The class distributions studied are balanced classes, 10\% and 5\% cut ins. In order to validate the results, all experiments are averaged over 10 runs and the faded colored lines represent the variance. We use the F1 score as the evaluation metric, since it is suitable for data with imbalanced classes. The F1 score is the harmonic mean between precision and recall, and is one of the most commonly used metrics. Unknown class detection is performed using the dataset with 10\% cut ins, regarding the cut in class as the unknown class.

\subsection{Investigation of embeddings for active learning}

\begin{figure}[htb]
    \centering
 \subfigure[SVM + Entropy]{\label{subfig:embeggings_svm_entropy}\includegraphics[width=0.3\textwidth]{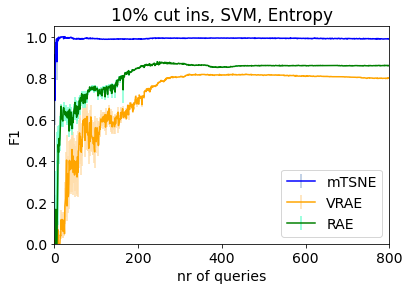}}
    \subfigure[SVM + Margin]{\label{subfig:embeggings_svm_margin}\includegraphics[width=0.3\textwidth]{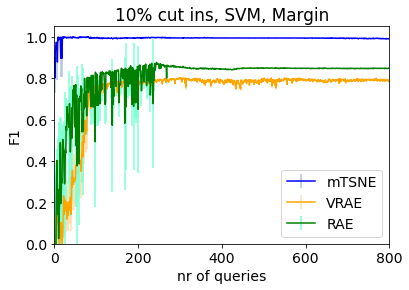}}
    \subfigure[SVM + Random]{\label{subfig:embeggings_svm_random}\includegraphics[width=0.3\textwidth]{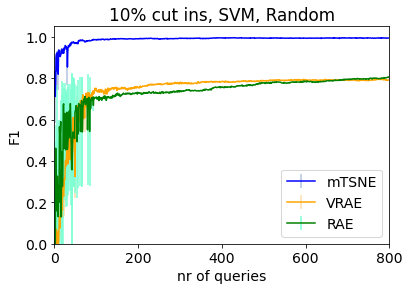}}
     \hfill
    \subfigure[NN + Entropy]{\label{subfig:embeggings_nn_entropy}\includegraphics[width=0.3\textwidth]{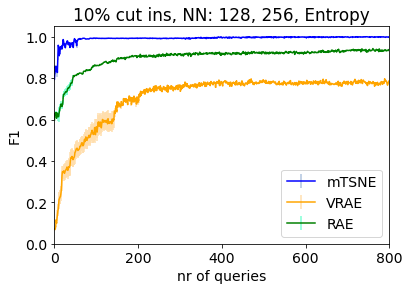}}
    \subfigure[NN + Margin]{\label{subfig:embeggings_nn_margin}\includegraphics[width=0.3\textwidth]{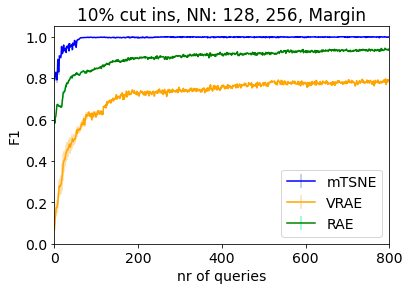}}
    \subfigure[NN + Random]{\label{subfig:embeggings_nn_random}\includegraphics[width=0.3\textwidth]{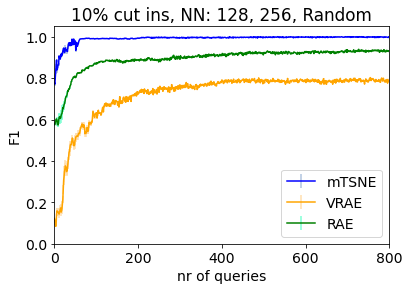}}
     \caption{Comparing the performance of mTSNE, RAE and VRAE in different settings.}
    \label{fig:embedding_comparison}
\end{figure}

Figure \ref{fig:embedding_comparison} shows the results when using different embeddings with SVM (first row) and NN (second row) as the classification model.
We observe that mTSNE achieves the highest F1 score. RAE also performs well in particular with NN.
 The VRAE embedding yields less promising results. The reason for why mTSNE outperforms the other embeddings is probably that its latent space is highly separable (as shown for example in Figure \ref{fig:mtsne_latent}), which makes the classification task easy.

The reason why VRAE performs worse could be explained simply by the nature of VRAE. Each trajectory is mapped to a distribution in the latent space with a larger error margin. This means that the margins between the classes are blurred and these points are a mixture of the two bordering classes. Such points might not give more information compared to a randomly drawn point and hence it becomes difficult to classify. Margin queries data points that have a high likelihood to belong to two different classes, and entropy queries those with the highest uncertainty of belonging to the most certain class. This would indicate that both margin and entropy query the data points that likely fall between the classes.

\subsection{Investigation of query strategies}
\begin{figure}[htb!]
    \centering
 \subfigure[mTSNE + SVM, $\alpha = 10$]{\label{subfig:QS_mtsne_svm}
    \includegraphics[width=0.45\textwidth]{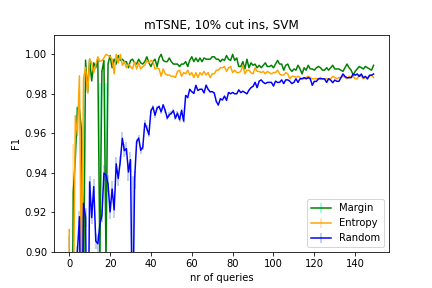}}
    \subfigure[mTSNE + SVM, $\alpha = 33$]{\label{subfig:QS_mtsne_svm_balanced}
    \includegraphics[width=0.45\textwidth]{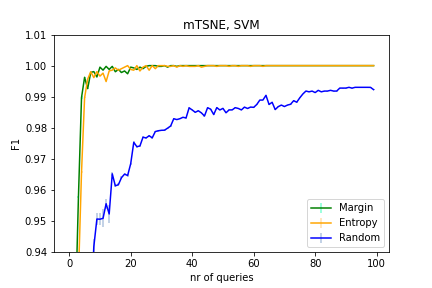}}
    \\
 \subfigure[RAE + SVM, $\alpha = 10$]{\label{subfig:QS_rae_svm}
    \includegraphics[width=0.45\textwidth]{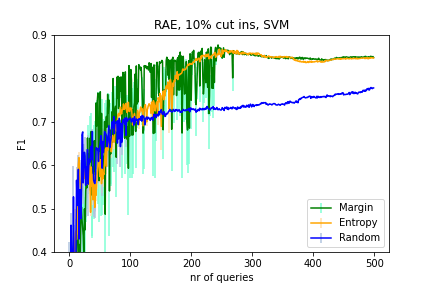}}
    \subfigure[RAE + SVM, $\alpha = 33$]{\label{subfig:QS_rae_svm_balanced}
    \includegraphics[width=0.45\textwidth]{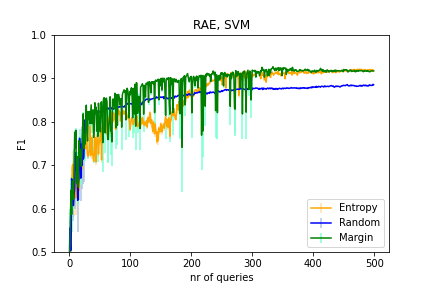}}
     \\
 \subfigure[VRAE + SVM, $\alpha = 10$]{\label{subfig:QS_vrae_svm}
    \includegraphics[width=0.45\textwidth]{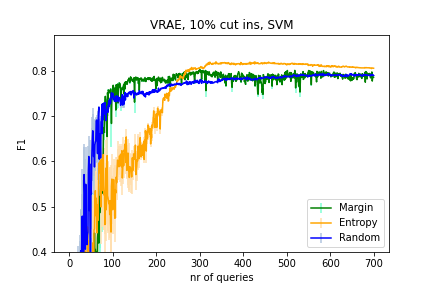}}
    \subfigure[VRAE + SVM, $\alpha = 33$]{\label{subfig:QS_vrae_svm_balanced}
    \includegraphics[width=0.45\textwidth]{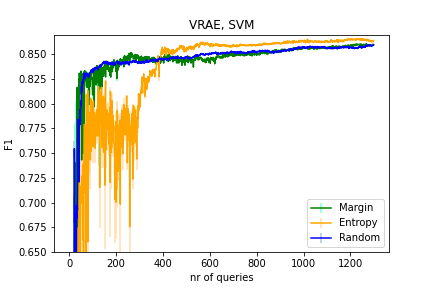}}
    \caption{The F1 score for the three query strategies using datasets with $\alpha = 33, 10$. We illustrate the results for each embedding, mTSNE (a)-(b), RAE (c)-(d) and VRAE (e)-(f).}
    \label{fig:QS}
\end{figure}

In this section, we study and compare the different query strategies.
Figure \ref{fig:QS} illustrates the three query strategies for different embeddings using the SVM classifier. In Figures \ref{subfig:QS_mtsne_svm}-\ref{subfig:QS_mtsne_svm_balanced}, margin and entropy outperform random achieving a high F1 score much faster, for the datasets with $\alpha = 10, 33$.

Figures \ref{subfig:QS_rae_svm}-\ref{subfig:QS_rae_svm_balanced} show the result for RAE embedded data. A similar behavior to mTSNE is observed, where using SVM the non-random query strategies (entropy and margin) yield better performance for both class distributions. In particular, entropy obtains a stable and high F1 score in this setting.
A general trend observed for the mTSNE and RAE embedded data is that SVM in combination with margin query tends to yield larger fluctuations, but with a stable baseline. This behavior is especially visible in Figure \ref{subfig:QS_rae_svm}, where margin fluctuates more than entropy. A possible reason could be that margin, as the name indicates, queries the data points with the smallest margin. This means that SVM is more sensitive w.r.t. the new data points queried by margin, as the separating hyperplane can change its direction rapidly with new data points.

Looking at Figure \ref{subfig:QS_vrae_svm}, it is clear that entropy does not perform well when using VRAE. We also observe that there is no big difference between margin and random. 
However, this embedding yields in overall lower performance compared to mTSNE and RAE.
As mentioned before, in this case the queried data points probably come from the boundary space  between the classes, and do not provide more information than a randomly queried data point.

\subsection{Investigation of choice of classifiers}
\begin{figure}[htb]
     \centering
 \subfigure[mTSNE + NN, $\alpha = 10$]{\label{subfig:classifier_nn_mstne}\includegraphics[width=0.36\textwidth]{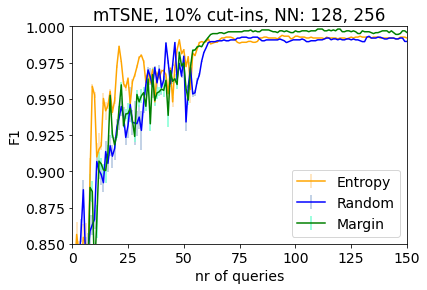}}
\subfigure[RAE + NN, $\alpha = 10$]{\label{subfig:classifier_nn_rae}\includegraphics[width=0.36\textwidth]{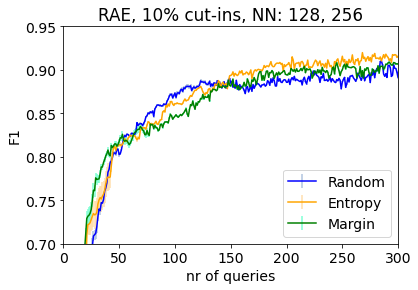}}
 \subfigure[VRAE + NN, $\alpha = 10$]{\label{subfig:classifier_nn_vrae}\includegraphics[width=0.36\textwidth]{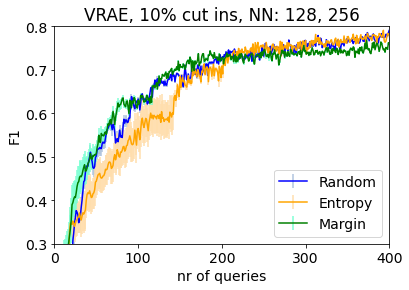}}
 \caption{The F1 score using NN classifier.}
    \label{fig:classifiers_nn}
\end{figure}

\begin{figure}
    \centering
\includegraphics[width = 0.6\textwidth]{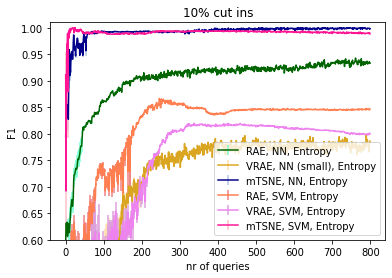}
    \caption{Comparing the F1 score using SVM and NN for each embedding.}
    \label{fig:classifiers_comparison}
\end{figure}

In this study, we investigate the choice of classification model for active learning in our datasets.
It has been observed that active learning is more effective in combination with SVM for all embeddings. This trend can be seen when comparing Figure \ref{fig:classifiers_nn} (which uses NN) to Figure \ref{fig:QS} (which uses SVM), where there is no significant difference in performance among the query strategies when we use an NN. For RAE, NN gives a higher F1 score than SVM, but for mTSNE and VRAE using SVM yields better performance compared to NN, as seen in Figure \ref{fig:classifiers_comparison}. However, SVM tends to show larger fluctuations than NN, because of higher sensitivity to newly queried data points.

The observation that the entropy and margin query strategies do not perform better in the case of using an NN, indicates the benefits from active learning do not contribute to the improved performance. This observation is consistent with the study in \cite{AL4Reaction2021} on active learning for chemical reaction prediction. This behavior seems to be due to the fact that an SVM can recognize general patterns fast, while an NN needs longer time to learn (due to its larger capacity). On the other hand, an NN is capable of learning more complex patterns and features.

Since the VRAE embedding yields overall worse results than the other embeddings, a larger NN consisting of five hidden layers with 64, 128, 256, 128 and 64 neurons is tested. Despite the increased capacity, the performance does not improve. There are several factors that could contribute to the stagnation at 0.8, such as the nature of VRAE and model configurations.
In general one can conclude that a simpler model like SVM is sufficient and suitable to be used in combination with active learning for our datasets.

\subsection{Investigation of class distributions}

\begin{figure}[htb]
\begin{adjustwidth}{-1cm}{-1.1cm}
     \centering
 \subfigure[mTSNE]{\label{subfig:class_distr_mtsne}\includegraphics[width=0.42\textwidth]{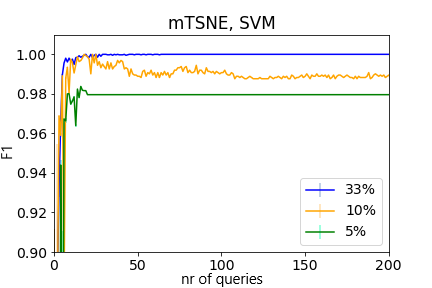}}
\subfigure[RAE]{\label{subfig:class_distr_rae}\includegraphics[width=0.42\textwidth]{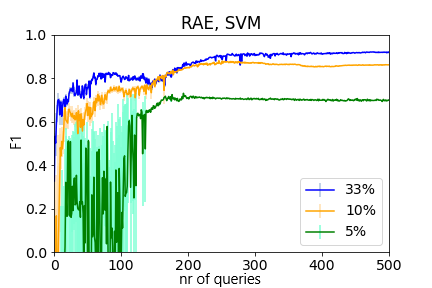}}
\subfigure[VRAE]{\label{subfig:class_distr_vrae}\includegraphics[width=0.42\textwidth]{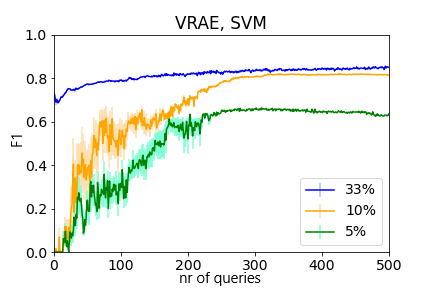}}
     \caption{Comparing the performance of class distributions containing 33\%, 10\% and 5\% cut ins. SVM is used with (a) mTSNE, (b) RAE and (c) VRAE embeddings.}
    \label{fig:class_distr}
\end{adjustwidth}
\end{figure}

In Figure \ref{fig:class_distr}, we investigate different class distributions with 33\%, 10\% and 5\% cut ins, and report the F1 scores for the three embeddings (we use SVM as the classifier). For all embeddings the class distribution with $\alpha$ = 33 seems to give the best results, as it saturates the fastest and obtains the highest F1 score. The dataset with $\alpha$ = 33 only performs slightly better than the one with $\alpha$ = 10 for mTSNE and RAE, while the difference is much larger for the VRAE embedding. Having $\alpha$ = 5 yields poor performance in all cases, since the cut in class is under represented.

Even though having an equal number of data points from each class might be expected to yield a better performance by large margins, it is still not a trivial matter. A higher percentage of cut ins leads to a larger variation as well, since cut ins can come in several forms difficult to recognize, such as double cut ins and decelerative cut ins.

\subsection{Investigation of budget size}

Figure \ref{subfig:class_distr_mtsne} demonstrates that with mTSNE, a high F1 score $\approx 1$ can be achieved within 25 queries for the different considered values of $\alpha$. With RAE and the SVM classifier, for $\alpha = 33, 10$ an F1 score of above $0.8$ is achieved after approximately 225 queries, and for $\alpha = 5$ an F1 score of around 0.7 is achieved after 200 queries; see Figure \ref{subfig:class_distr_rae}. Note that SVM is not the optimal choice of classifier for RAE. Looking at Figure \ref{subfig:classifier_nn_rae}, only around 125 queries is required to obtain an F1 score above 0.9 with RAE for $\alpha = 10$. Figure \ref{subfig:class_distr_vrae} shows that with VRAE, the F1 score plateaus around 0.8 and 0.6 after around 250 queries for $\alpha = 10, 5$ respectively. For $\alpha = 33$, the saturation at 0.8 occurs within 25 queries.

\begin{figure}[htb]
\begin{adjustwidth}{-1cm}{-1.1cm}
     \centering
     \subfigure[mTSNE]{\label{subfig:rate_f1_mtsne}\includegraphics[width=0.42\textwidth]{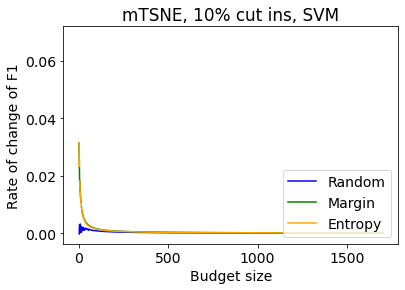}}
     \subfigure[RAE]{\label{subfig:rate_f1_rae}\includegraphics[width=0.42\textwidth]{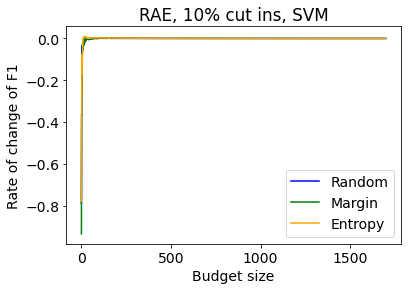}}
\subfigure[VRAE]{\label{subfig:rate_f1_vrae}\includegraphics[width=0.42\textwidth]{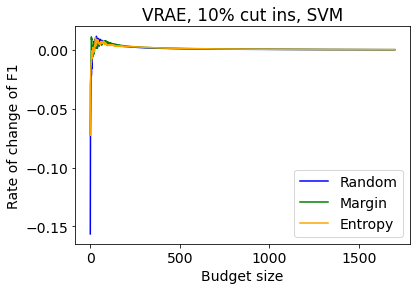}}
    \caption{The rate of change of F1 score for each embedding using the SVM classifier.}
    \label{fig:rate_of_f1}
\end{adjustwidth}
\end{figure}

Another way of looking at the optimal budget size is to investigate the rate of change of the F1 score shown in Figure \ref{fig:rate_of_f1}. Consistent with the other results, we observe that mTSNE with SVM is an optimal choice of set up. The rate of change of F1 score reaches its plateau with the smallest budget size and minimal fluctuations. On the other hand, it is the set up that exhibits a meaningful behavior in the sense that the informativeness rate decreases as more data points are queried.

\subsection{Investigation of unknown class detection}

\begin{figure}[ht!]
     \centering
\subfigure[mTSNE + SVM]{\label{subfig:unknown_mtsne_svm_200}
     \includegraphics[width=0.42\textwidth]{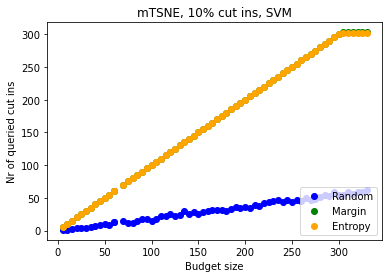}}
     \subfigure[mTSNE + NN]{\label{subfig:unknown_mtsne_nn_200}
     \includegraphics[width=0.42\textwidth]{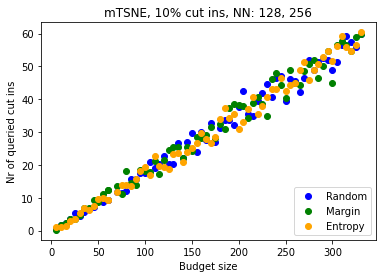}}
 \subfigure[RAE + SVM]{\label{subfig:unknown_rae_svm_200}
     \includegraphics[width=0.42\textwidth]{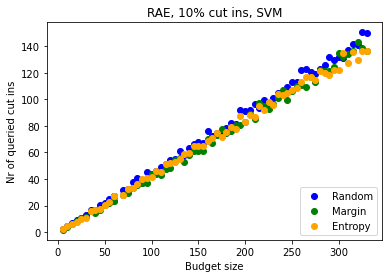}}
     \subfigure[VRAE + SVM]{\label{subfig:unknown_vrae_svm_200}
     \includegraphics[width=0.42\textwidth]{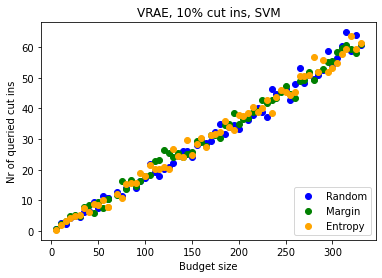}}
      \\
\subfigure[RAE + TSNE + SVM]{\label{subfig:unknown_rae_tsne_svm}
     \includegraphics[width=0.42\textwidth]{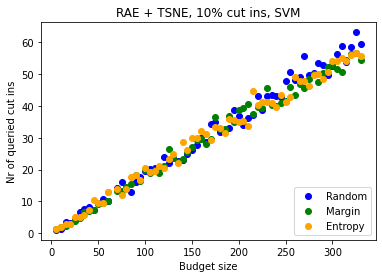}}
     \subfigure[VRAE + TSNE + SVM]{\label{subfig:unknown_vrae_tsne_svm}
     \includegraphics[width=0.42\textwidth]{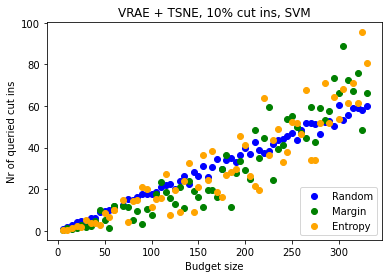}}
     \caption{Number of cut ins queried against budget size during 330 queries.}
    \label{fig:unknown}
\end{figure}

\begin{figure}[ht!]
     \centering
 \subfigure[Margin]{\label{subfig:embedding_mtsne_mrg}\includegraphics[width=0.42\textwidth]{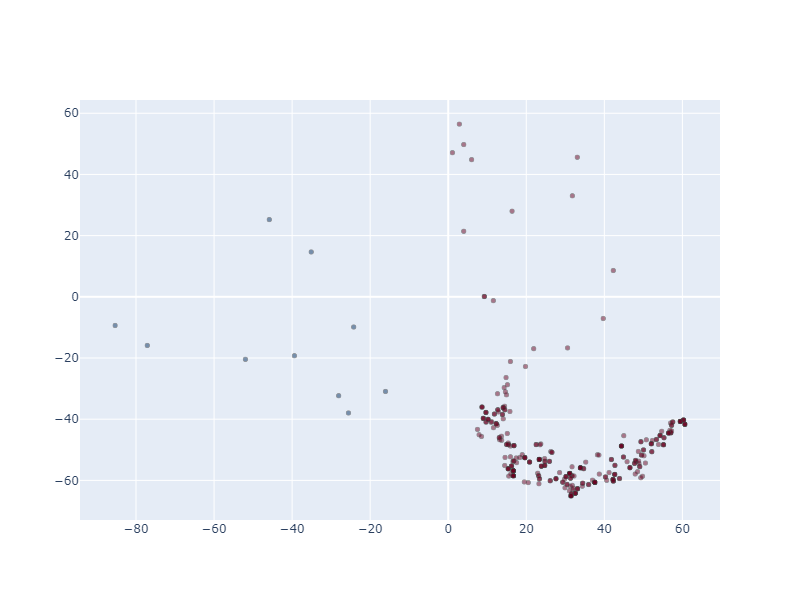}}
\subfigure[Entropy]{\label{subfig:embedding_mtsne_entr}\includegraphics[width=0.42\textwidth]{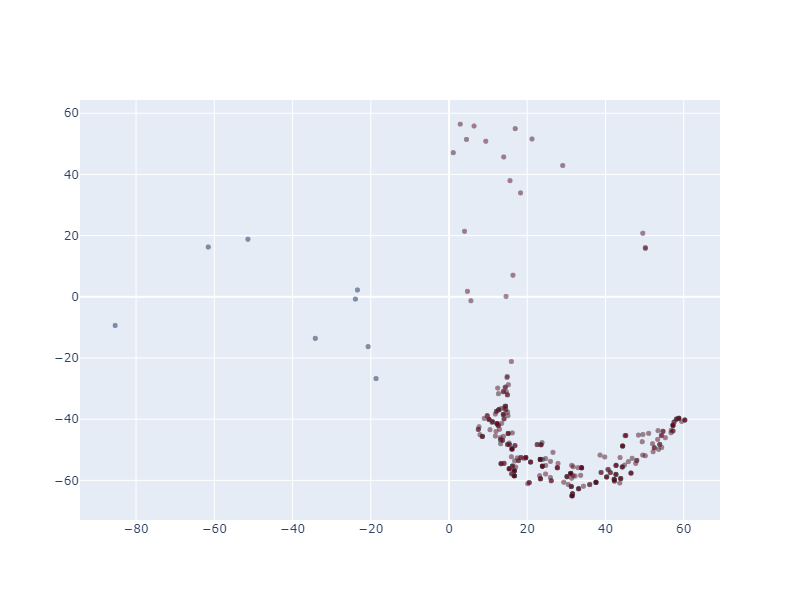}}
 \subfigure[Random]{\label{subfig:embedding_mtsne_rnd}\includegraphics[width=0.42\textwidth]{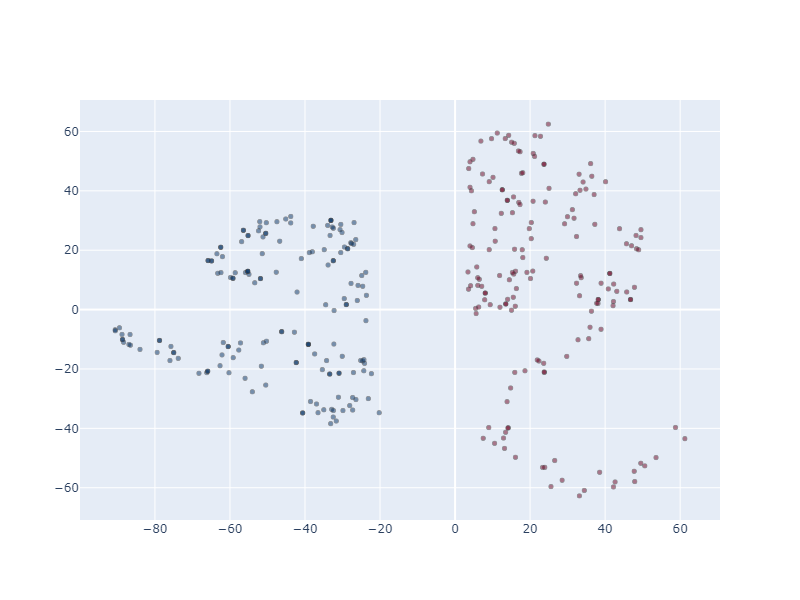}}
    \caption{The queried trajectories in mTSNE latent space after 330 queries, with (a) margin, (b) entropy and (c) random using SVM.}
    \label{fig:mtsne_queried_points}
\end{figure}

Finally, we investigate unknown class detection using the proposed active learning framework, where the cut in class is considered as the unknown class. Figure \ref{fig:unknown} shows the number of queried cut ins for each embedding over 330 queries, using SVM or NN as the classification models. Figure \ref{subfig:unknown_mtsne_svm_200} and \ref{subfig:unknown_mtsne_nn_200} show the results for the mTSNE embedded data. The results with SVM are more impressive compared to NN (therefore, for the two other embeddings we focus only on SVM). In Figure \ref{fig:mtsne_queried_points}, we illustrate the queried points displayed in mTSNE latent space.
We observe that the entropy and margin strategies query more cut ins using SVM.
With the NN classifier, the  entropy and margin strategies perform almost as good as the random choice.
The reason could be, as mentioned earlier, SVM is a simpler classification model which learns and saturates faster compared to NN (which is a model with a large capacity). Thus, SVM quickly learns the two existing classes (left and right drive by) and then starts querying the unknown class members. On the other hand, NN needs a lot of data for a proper learning and continuously needs data points from left and right drive by classes.

Figures  \ref{subfig:unknown_rae_svm_200} and \ref{subfig:unknown_vrae_svm_200} show the unknown class  detection results for RAE and VRAE embeddings employed with SVM. Figures \ref{subfig:unknown_rae_tsne_svm} and \ref{subfig:unknown_vrae_tsne_svm} show the results for RAE and VRAE embedded with TSNE using SVM, which could improve the results further compared to RAE and VRAE without TSNE. We observe that in both settings the entropy or margin querying strategies can be helpful for identifying the cut ins. This is even more obvious when using the VRAE combined with TSNE embedding.

\section{Conclusion}
In this study, we investigated the performance of active learning as an effective tool for reliable and cost-efficient labeling of the time series trajectory data collected from Autonomous Drive (AD) application. For this purpose, we developed a framework wherein we first embed the trajectories into a latent space representation (for example using mTSNE, RAE and VRAE) in order to extract the temporal nature of the trajectories. We then apply the active learning paradigms using different querying strategies and classification models in the embedded latent space.
We  also explored the possibilities for unknown class detection using the proposed active learning framework.

We observe that in many settings, active learning constitutes an effective tool. The positive effect is particularly more obvious with the SVM classifier, for both  mTSNE and RAE embeddings. The class distribution yielding the best performance is when $\alpha$ = 33, that is only slightly better than $\alpha$ = 10. The choice of a proper embedding affects significantly the results. In particular, mTSNE yields consistently the best performance compared to the alternatives. With mTSNE, the entropy querying strategy used with SVM can be seen as the best option due to high performance with a small number of queries, and a higher stability than margin.

RAE still performs well in particular compared to VRAE.
With the RAE embedding, using an NN yields better performance than SVM, with no explicitly significant difference in performance among the query strategies. The VRAE embedding does not achieve a particularly high performance regardless of the choice of the classifier. There is no significant difference between random and margin, but entropy performs somewhat worse in this setting.

Regarding unknown class detection, we observe that the proposed active learning framework can be useful for this task as well. In particular, when we use SVM with mTSNE embedding, we see that the entropy or margin query strategies yield identifying more cut ins compared to the random strategy. Due to large capacity, this is less obvious with NN.

\section*{Acknowledgement}
We would like to acknowledge Volvo Cars Corporation for providing the data and computational resources. The work of Linus Aronsson and Morteza Haghir Chehreghani was partially supported by the Wallenberg AI, Autonomous Systems and Software Program (WASP) funded by the Knut and Alice Wallenberg Foundation.

\bibliographystyle{plain}
\bibliography{ref,ref_new}

\end{document}